\documentclass[10pt,twocolumn,letterpaper]{article}
\pdfoutput=1

\usepackage{cvpr}
\usepackage{times}
\usepackage{epsfig}
\usepackage{graphicx}
\usepackage{amsmath}
\usepackage{amssymb}
\usepackage{color}

\usepackage{threeparttable}
\usepackage{multirow}

\usepackage[pagebackref=true,breaklinks=true,letterpaper=true,colorlinks,bookmarks=false]{hyperref}

\cvprfinalcopy 

\def\cvprPaperID{0} 

\newcommand{\denselist}{\itemsep 0pt\parsep=0pt\partopsep0pt\vspace{-\topsep}}

\newcommand{\bitem}{\begin{itemize}\denselist}
	\newcommand{\eitem}{\end{itemize}}
\newcommand{\benum}{\begin{enumerate}\denselist}
	\newcommand{\eenum}{\end{enumerate}}
\newcommand{\bdescr}{\begin{description}\denselist}
	\newcommand{\edescr}{\end{description}}

\ifcvprfinal\fi

\begin{document}

\title{View Extrapolation of Human Body from a Single Image}

\author{
Hao Zhu$^{1,2}$ \quad Hao Su$^{3}$ \quad Peng Wang$^{4,5}$ \quad Xun Cao$^{1}$ \quad Ruigang Yang$^{2,4,5}$\\
$^{1}$Nanjing University, Nanjing, China \quad $^{2}$University of Kentucky, Lexington, KY, USA\\
$^{3}$University of California, San Diego, CA, USA \quad $^{4}$Baidu Inc., Beijing, China\\
$^{5}$National Engineering Laboratory of Deep Learning and Technology and Application, China\\
\tt\small{zhuhao\_nju@163.com \quad haosu@eng.ucsd.edu \quad \{wangpeng54, yangruigang\}@baidu.com \quad caoxun@nju.edu.cn}
}

\maketitle

\begin{abstract}
We study how to synthesize novel views of human body from a single image. Though recent deep learning based methods work well for rigid objects, they often fail on objects with large articulation, like human bodies. The core step of existing methods is to fit a map from the observable views to novel views by CNNs; however, the rich articulation modes of human body make it rather challenging for CNNs to memorize and interpolate the data well. To address the problem, we propose a novel deep learning based pipeline that explicitly estimates and leverages the geometry of the underlying human body. Our new pipeline is a composition of a shape estimation network and an image generation network, and at the interface a perspective transformation is applied to generate a forward flow for pixel value transportation. Our design is able to factor out the space of data variation and makes learning at each step much easier. Empirically, we show that the performance for pose-varying objects can be improved dramatically. Our method can also be applied on real data captured by 3D sensors, and the flow generated by our methods can be used for generating high quality results in higher resolution.
\end{abstract}

\section{Introduction}
\label{sec:intro}

In recent years, a number of papers have been published on inferring 3D structures from single images using learning based approaches \cite{CVPR2015flynn, NIPS2015Yang, ECCV2016Garg, ECCV2016zhou, CVPR2017park, CVPR2017ji, ICCV2015Su, TOG2016Kalantari, NIPS2014Eigen, SIGGRAPH2014Su, TCSVT2017Cao, ICRA2018Ma, 3DV2016Laina, CVPR2017ZFan}. One important task in this topic is \emph{novel-view synthesis} - predicting what a given object would look like after a known 3D rotation is applied. In psychology, this task is known as ``mental rotation'' and experiments tell us that people excel at this task \cite{SCIENCE1971NShepard}.  Practically, addressing this problem would also have far-reaching impacts in image editing, augmented reality, virtual reality, and many other applications.

\begin{figure}[t]
\begin{center}
   \includegraphics[width=1.0\linewidth]{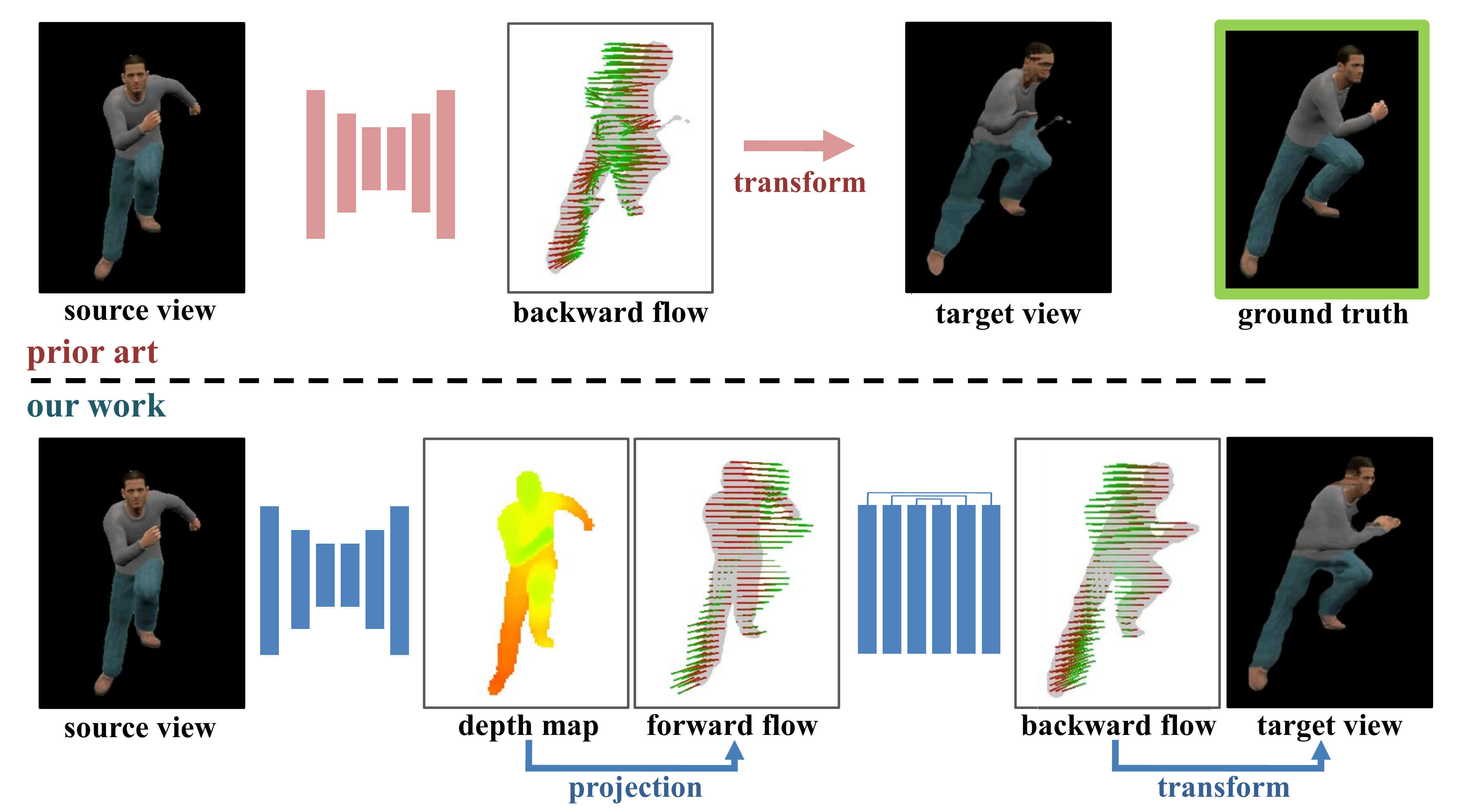}
\vspace{-2em}
\end{center}
   \caption{When dealing with pose-varying human body, previous methods like VSAP\cite{ECCV2016zhou} fail to predict accurate flow. In contrast to VSAP that directly predicts 2D flow, our method firstly predicts the depth map, then the forward flow based framework dramatically improve the accuracy of the flow prediction.  This architecture reflects our \emph{appearance - shape - flow} strategy.}
\label{fig:Fig1_glimpse}
\vspace{-1em}
\end{figure}

In principle, inferring 3D geometry from a single image is an ill-defined problem. 
Recently, \cite{ICCV2015Su, TPAMI2017Alex, ECCV2016zhou, CVPR2017park, CVPR2017ji, NIPS2015Yang} have shown that it is quite promising to learn a cross-view synthesizer from many source/target view pairs. 

While decent results have been obtained for rigid objects like vehicles, existing approaches perform quite poorly on human bodies, which are both articulated and deformable. As can be seen in Figure \ref{fig:Fig1_glimpse}, the prediction from previous methods is often blurry or distorted. To understand why, let's take a closer look into these approaches. Typically, they either directly predict each pixel in the target view \cite{ECCV2016tatarchenko, CVPR2017park, NIPS2015Yang, ICCV2017Huang, arXiv2017Zhao}, or predict a flow map that represents where pixels should be copied from in the source view (backward flow prediction)\cite{ECCV2016zhou, CVPR2017ji, ECCV2016Garg, NIPS2015jaderberg, ICANN2011Hinton}. Compared with rigid objects, the articulation of moving parts of humans, like limbs and heads, may differ greatly from case to case. Coupled with the orientation of human body, part deformation and source/target view, this kind of pose variation forms an even larger joint space for all possible source/target sample pairs.  Therefore, it becomes extremely difficult for the network to remember and interpolate such highly varied data from limited training samples. 

Our approach, based upon the flow prediction idea, is a combination of deep learning and traditional geometry-based methods.  Instead of resorting to purely learning based schemes, we consider how to explicitly leverage geometric constraints to reduce the problem space.  The basis of our idea is that the flow across two views can be analytically derived if corresponding 3D geometry is known. 

We therefore decouple the flow prediction network into a concatenation of shape estimation sub-network and an image generation sub-network, each supervised separately with additional labels. Intuitively speaking, the first sub-network estimates a rough geometry of the human body and the second sub-network corrects the error caused by the inaccuracy and infers invisible regions caused by occlusions.  In this way, the entanglement between shape estimation, flow estimation, and invisible region synthesis are detached.  If we take a geometric perspective, the high-dimensional space that data samples live in has been factored out into lower-dimensional subspaces. Between the two sub-networks, perspective projection is applied to propose a flow. It turns out that, this well-defined perspective projection is difficult to be learned implicitly by traditional convolutional neural network according to our experiments, which is another observation to support our design.  Empirically, our hybrid approach generates more accurate results on standard benchmarks, compared with state-of-the-arts. 

The contributions of this paper include:
\bitem
\item We develop a novel approach that firstly predicts object shape from appearance, then predicts optical flow from the shape. The flow loss and image loss are integrated to improve the prediction accuracy while the structure in novel view is maintained. By using an explicit 3D representation, we are able to handle large deformation in shape and large changes in view direction. 
\item We have empirically found out that traditional convolutional neural network cannot fit per-pixel projection well. Combining geometric projection with CNN is more effective than an end-to-end CNN-based approach. 
\item We have created a high-quality data sets for synthetic human body images with over 2,000 different poses and 22 different appearances. The data set has been publicly released\footnote{\url{http://cite.nju.edu.cn/view_extrapolation.html}}.  
\eitem

\vspace{0.1in}
\section{Related Work}
\label{sec:related}

\noindent\textbf{View Synthesis by 3D modeling. }  Traditional view synthesis approaches follow the \emph{modeling - rendering} pattern, which take advantage of 3D reconstruction method\cite{CVPR2006Seitz, TPAMI2010Furukawa, FCS2017Zhu, CVPR2019Liu, TCSVT2017Zhu} to generate 3D model, then render the target image in novel views.  These methods require high-quality 3D shapes in forms of polygen meshes, point clouds, or depth images.  It is beyond the scope of this paper to discuss the vast body of 3D reconstruction and view synthesis techniques.  We will focus on single-image-based techniques here.  Given the reference 3D model sets, Kholgade et al. \cite{TOG2014Kholgade} generated the synthesis images by manually interactive manipulating the existing 3d model sets.  Su et al.\cite{ICCV2015Su} proposed to synthesize new features for other views of the same object by finding the correspondence patches from an aligned set of 3D models.  Rematas et al. \cite{TPAMI2017Rematas} proposed to fit the shape in the 3D data sets to the source image, then synthesize the high-quality image in the novel view.  These methods require the aid of certain categories of 3D shapes prior.  

For human pose/body modeling, the generative SCAPE model~\cite{SCAPE:sig05} has been widely used. This model and its variants have been used to estimate 3D body shape and/or pose from a single image (e.g.,~\cite{Sigal:nips:2008,Bogo:ECCV:2016}). However, there is no variations in skin appearance in the these statistical models, and none of these works address the view synthesis either. Given the visible misalignment between the image and the 3D model, it is unlikely that a direct rendering of the 3D model is able to generate satisfactory results.

Recently the introduction of learning based methods makes it possible to predict the shape from a single image ~\cite{IJCV2008Saxena, NIPS2014Eigen, SIGGRAPH2014Su, CVPR2017ZFan}.  However, low granularity of the predicted shape limits their application in image synthesis.  Another group of work focuses on incorporating shape information inside the neural network.  Flynn et al. \cite{CVPR2015flynn} proposed to turn plane sweep stereo into label prediction problem, which proceeds depth prediction and view synthesis in an end-to-end convolutional neural network.  Garg et al. \cite{ECCV2016Garg} proposed to predict the depth from the single image in an unsupervised manner.  Their network warps the source image to the target image in training phase, and explicitly predicts an image from one of the stereo pairs to the other.  Srinivasan et al. \cite{ICCV2017Srinivasan} proposed to synthesize the 4D light field by first predicting ray depth, and then rendering a Lambertian approximation to the light field.  These two methods explicitly take advantage of depth or disparity information in their network.  However, they focus on relatively short base-line view synthesis, while our system could be used in wide base-line views.  Besides, the pose-varying human body is more challenging compared with static scenes as the human body is articulated non-rigid.

\begin{figure*}
\begin{center}
\includegraphics[width=1\linewidth]{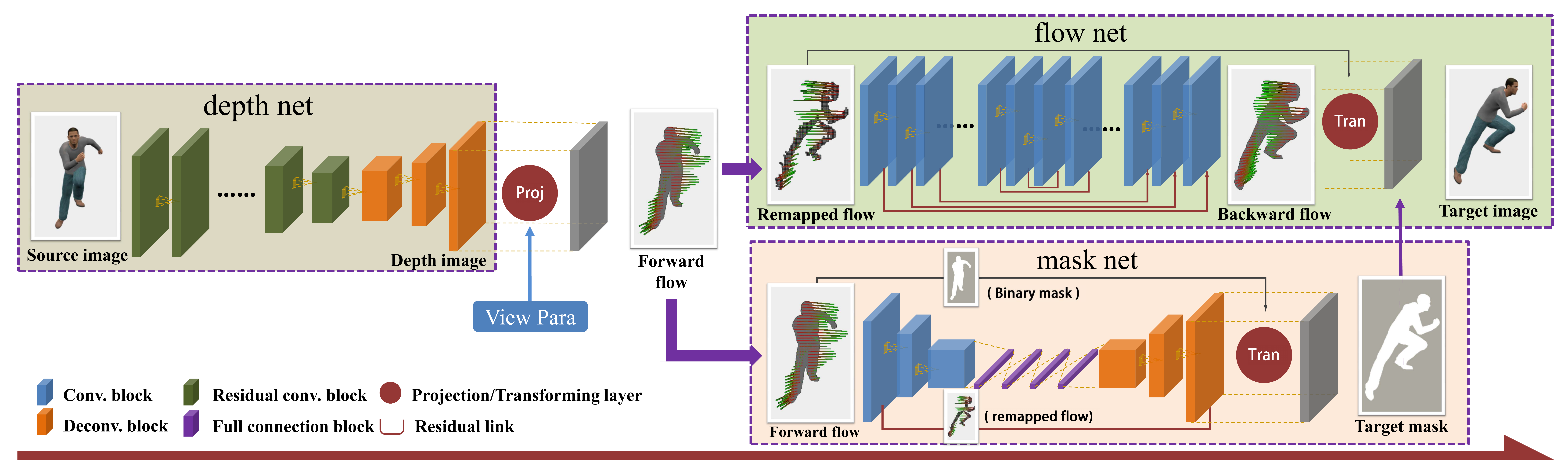}
\vspace{-2.5em}
\end{center}
   \caption{The full pipeline of our approach.  The architecture of network is simplified, and the detailed parameters will be shown in supplement materials.}
\label{fig:Fig2_pipeline}
\vspace{-1em}
\end{figure*}

\vspace{0.1in}
\noindent\textbf{View Synthesis by 2D Flow. }  The core idea behind flow is to estimate the pixel mapping from the source image to the target image.  Therefore, these approaches aim to learn to produce a flow image or a set of mapping correspondences to transform source image to target image.  The transforming auto-encoders \cite{ICANN2011Hinton} and the spatial transformer networks \cite{NIPS2015jaderberg} firstly implement this flow estimation process into the neural networks, where the spatial transforming map is learned from source image and then guides target image synthesis.  Zhou et al. \cite{ECCV2016zhou} proposed to synthesize from one view to another view of one certain category of objects by predicting optical flow, namely appearance flow.  Their method achieves good results in certain categories of objects like cars and chairs. 

\noindent\textbf{View Synthesis by Direct Image Generation.  }  The previous works presented by Tatarchenko et al. \cite{ECCV2016tatarchenko} and Yang et. al \cite{NIPS2015Yang} solved the view synthesis problem by implementing a convolutional neural network to directly predict target images.  Their methods cannot preserve the details, which is a common problem for direct image generation network.  To generate high-quality images, novel techniques like generative adversarial nets (GANs)~\cite{NIPS2014Goodfellow} and perceptual loss~\cite{ECCV2016Johnson} are involved to tackle view synthesis problems.  Park et al. \cite{CVPR2017park} proposed to infer the invisible part on the base of appearance flow net \cite{ECCV2016zhou}, where an image generation network using perceptual and adversarial loss is supplemented to complete the invisible region.  Huang et al. \cite{ICCV2017Huang} proposed a GAN based view synthesis system aiming at high-quality face rotation.  The generator in their network consists of a local pathway and a global pathway, which detects feature layout and global appearance respectively.  Zhao et al. \cite{arXiv2017Zhao} proposed a GAN in coarse-to-fine pattern to synthesize high-resolution results.  Their network aims at multi-view cloth images from a single view image regardless of the pose variation.  Chen et al. \cite{ICCV2017Chen} proposed a cascaded architecture that predicts high-resolution street images from semantic layouts.

Unlike all the methods above, we aim to synthesize novel views from a single image of human body.  We find both state-of-art 2D transforming and direct image generation methods failed to synthesize the fine result for the human body.  Our approach is based on 2D flow morphing methods, but we incorporate an explicit 3D model and geometric constraints to provide accurate flow to handle the large pose variation and view-point variations. 

\section{Method}

We propose an \emph{appearance - shape - flow} strategy for synthesizing novel views of pose-varying objects.  As demonstrated in Figure \ref{fig:Fig2_pipeline}, we first predict the shape as a depth map from the source image. Through the projection layer, this depth map is transformed to a forward flow image. Then, the flow net and mask net will predict an optical flow and a mask simultaneously. Finally, the flow image and mask image are combined to produce the final synthesized images.  We will explain module by module in the following sections.

\subsection{Depth Prediction Net}
We describe the shape of a human body in the form of depth images.  
A large body of methods have been proposed to predict depth from the monocular intensity image \cite{TCSVT2017Cao, 3DV2016Laina, ICRA2018Ma, ICRA2017Liao}.  We select Res-Net~\cite{CVPR02016He} as the backbone of our depth image prediction net.  The depth prediction net consists of an encoder and a decoder that is adapted from the standard ResNet-50 net.  
The L1 loss is adopted while pixels in the background region are omitted.  We mask out the final predicted depth using the silhouette in the source image.

\noindent\textbf{Shape vs Appearance. }  Images are formed by projecting a 3D shape to a 2D plane. Along the projection, the majority of shape information is lost due to the lack of depth dimension.  Prior 2D transformation based network directly predicts an optical flow from the image supervised by the 2D appearance at the target view.  When dealing with rigid objects like cars and chairs, the appearance flow as in \cite{ECCV2016zhou} can be directly predicted from images.  However, if the object is articulated and deformable, we find that such direct prediction often fails.

In fact, given a depth image, the flow can be directly computed by perspective transformation.  Therefore, we borrow the idea of traditional synthesis methods by 3D modeling, which explicitly produces depth as the intermediate result.  Our experiments show that the \emph{appearance - shape - flow} strategy outperforms the \emph{appearance - flow} strategy in the following aspects:
(1) The flow prediction accuracy is improved dramatically;
(2) Better performance is achieved when self-occlusion occurs, which is common for pose-varying human;
(3) Our system is more robust when applied to real human subjects captured by 3D sensors.

\noindent\textbf{Projection Layer}
To combine the depth prediction net and flow/mask prediction net, we propose the projection layer which transforms the depth image to an optical flow.  Projection and inverse projection are the essential steps for the transformation between 3D shapes and 2D images.  Alternatively, one may adopt convolutional neural networks to learn the projection transformation. However, we find that neither deep nor shallow convolutional neural networks is able to fit per-pixel projection calculation accurately.  
In our opinion, the reason why CNN doesn't work is that the projection involves complicated per-pixel matrix calculation and homogeneous coordinates normalization, which are difficult for CNN to fit.  Therefore, we build specific layers to analytically compute projection and inverse projection operations.

The projection layer takes depth image and the camera matrix in synthesis view as input to generate the forward flow.  The projection layer does not include any parameter to learn.  Given the depth image $D(u,v)$, intrinsic matrix $K$, and relative extrinsic matrix $Rt$ from target view to source view, the projection layer generates the forward flow image $flow(x, y)$ by the following calculations:

\vspace{-0.07in}
\begin{equation}
\label{eq1}
p = [ \frac{(x-c_x)}{f_x}, \frac{(y-c_y)}{f_y}, 1]^T \cdot D(x,y)
\end{equation}

\vspace{-0.07in}
\begin{equation}
\label{eq2}
\widetilde{uv} = K * Rt' \cdot \tilde{p}
\end{equation}

\vspace{-0.17in}
\begin{equation}
\label{eq3}
flow(x, y) = norm(\widetilde{uv}) - [x, y]^T
\end{equation}
where $p$ denotes the point in 3D world coordinates of source view.  $\widetilde{uv}$ is the coordinates of $uv$, and $norm()$ computes a point coordinate from the homogeneous vector. The intrinsic and extrinsic parameters are set according to the image rendering setup, and keep fixed in all our experiments. The output of the projection layer is the forward flow map, which will be explained in the next section.

\subsection{Flow Prediction Net}
We base our work on the idea of predicting the flow to transport pixel value across views. VSAP\cite{ECCV2016zhou} predicts a backward flow map to denote where pixels in the target image should be copied from in the source image, and then produce the target image by remapping the source image using the flow map.  Different from previous works, our network takes forward flow instead of intensity image as the input.

\begin{figure}[t]
\begin{center}
\includegraphics[width=1\linewidth]{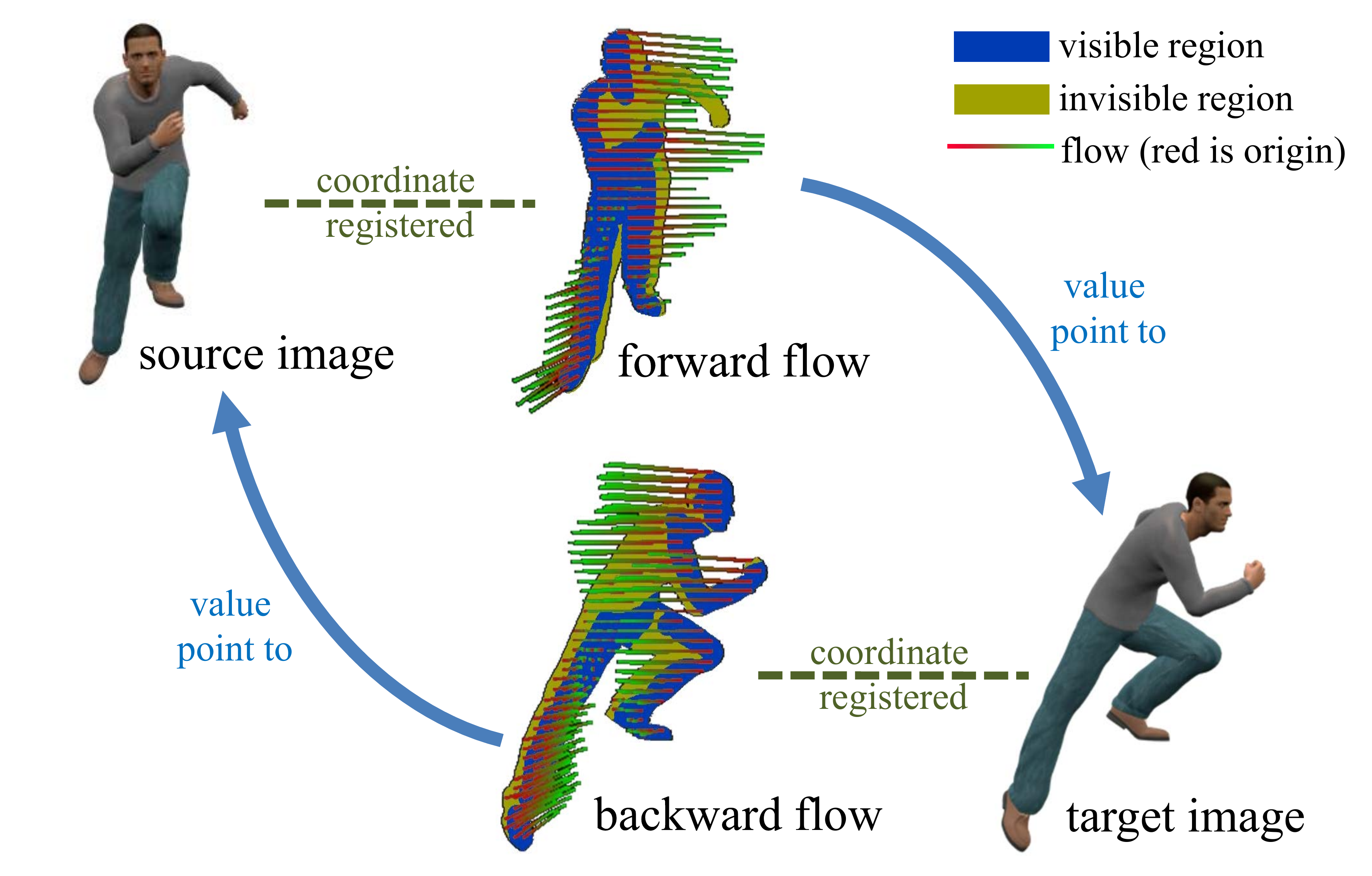}
\vspace{-2.5em}
\end{center}
   \caption{The forward flow and backward flow.}
\label{fig:Fig3_forward_flow}
\vspace{-1.5em}
\end{figure}

\noindent\textbf{Forward flow. }
First let us formally define the forward flow and backward flow.  Both forward flow and backward flow are two channel floating point arrays. Here, `forward' means a flow that denotes the pixel-wise correspondences from source image to target image, and `backward' means the reverse orientation of correspondences.  As shown in Figure \ref{fig:Fig3_forward_flow}, the forward flow is registered to source image, and the value in each pixel is the coordinate of the corresponding point in the target view image, $[u,v]$.  The backward flow is defined the other way around.

To convert forward flow and backward flow from one to the other, we have to consider the occlusion problem.  As shown in Figure \ref{fig:Fig3_forward_flow}, the dark yellow regions in forward flow or backward flow are invisible to the other, while only the blue part which are visible in both target view and source view can be transformed by copying the coordinate and value of every pixel.  In our forward-to-backward transformation process, we call the transformation result as transformed flow.  If one coordinate in the transformed flow is allocated with more than one value from the source image, we will select the value whose corresponding depth to the target view is the smallest.  In practice, given the target view coordinate, we set four closest neighboring pixels in the transformed flow to be the source coordinate.  So after the projection layer, the flow occlusion problem in the source view will be solved.  However, there will be blank regions in the transformed flow map that correspond to the invisible region in the target view.  We will complete the transformed flow in the flow prediction net.  

Previous works \cite{NIPS2015jaderberg, ICANN2011Hinton, ECCV2016zhou} choose to directly predict the backward flow.  However, we find that in our problem of view extrapolation for human bodies, it is better to generate the forward flow first. Compared with backward flow prediction, forward flow will benefits from two aspects: (1) The coordinate of the forward flow is registered to the source, thus the forward flow can be transformed from depth image and camera parameters without ambiguity.  (2) The introduction of forward flow separates and simplifies the occlusion problem. In this way, the depth prediction net will not need to address occlusion problem.  When the forward flow is generated from the projection layer, the occlusion can be resolved with z-buffering.  As shown in the next part, the flow net merely needs to focus on completing those missing regions and refining the flow.  According to our experiment, we find that this strategy significantly improves the flow predicting accuracy, especially in complicated regions with self-occlusion.

\noindent\textbf{Net architecture }
As discussed above, the aim of the flow net is no longer to extract the flow from the source, but to refine the transformed flow to predict the final backward flow and image. As pointed out by a few recent papers \cite{ECCV2016zhou,CVPR2017park,NIPS2015Yang,ECCV2016tatarchenko}, traditional encoder-decoder network may lose details in the source image and thus generate blurry images. We also observed the same problem appearing in flow prediction.  We select the image restoration network~\cite{NIPS2017Mao} as the backbone structure to address our backward flow completion problem. Experiments show that this architecture efficiently restores the blank region in the flow with better preservation of details.

We propose to integrate image loss and flow loss in the flow net, as image loss alone is often affected by texture ambiguity.  The loss function is formulated as:

\vspace{-0.2in}
\begin{equation}
loss = \eta_y\sum_{i\in M}^{m}|y^{(i)}-\widehat{y}^{(i)}| + \eta_f\sum_{i\in N}^{n}|f^{(i)}-\widehat{f}^{(i)}|
\label{equ:loss_func}
\end{equation}
\vspace{-0.12in}

where $y^{(i)}$ is the RGB image and $f^{(i)}$ is the masked ground truth flow.  $\widehat{y}^{(i)}$ and $\widehat{f}^{(i)}$ are the corresponding ground truth image and backward flow used in training phase.  $\eta_y$ and $\eta_f$ are the the weight of image loss term and flow loss term separately.  $M$ is the pixel set including all foreground pixels.  $N$ is the pixel set including all pixels with valid backward flow, which means the invisible pixels at the source view will not count.  The ground truth backward flow is generated by projecting the 3D vertices movement to the target view.  
Our experiments show that the `image loss + flow loss' improves the prediction accuracy compared with `single image loss'.

\subsection{Mask Prediction Net}

\begin{figure}[t]
\begin{center}
\includegraphics[width=1\linewidth]{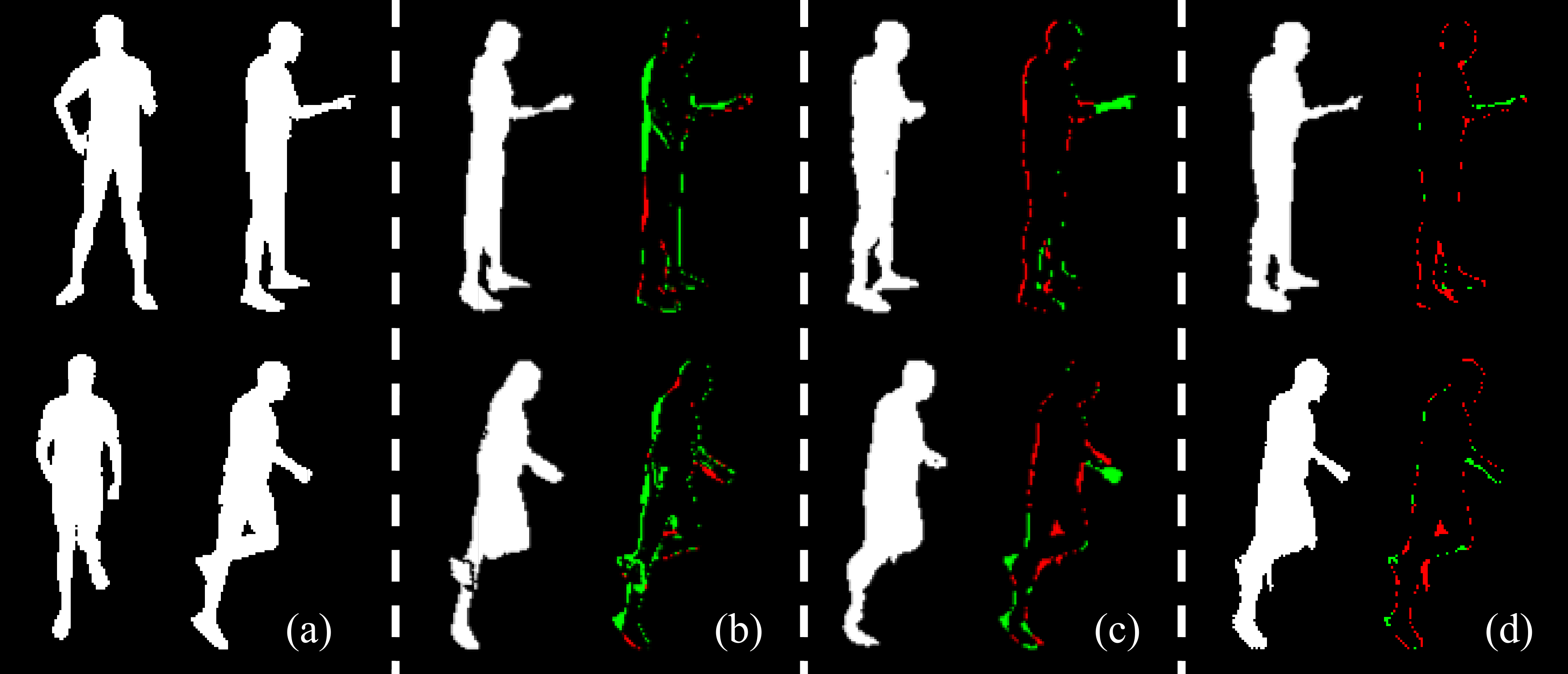}
\vspace{-2.5em}
\end{center}
   \caption{The result of different mask prediction network and corresponding error image. Green indicates that the algorithm shrinks compared with ground truth, and red means the prediction exceed ground truth.  The contents from left to right are: (a) Source / target image;  (b) Image restoration net; (c) Encode-decode net; (d) Residual mask prediction on the base of (c).}
\label{fig:Fig4_mask}
\vspace{-1em}
\end{figure}

The mask prediction net produces a silhouette of the object in the target view.  We follow the appearance flow network\cite{ECCV2016zhou} that predicts mask and flow simultaneously, and then fuse them to build the final predicted image.  In their network, the mask is directly predicted using an encoder-decoder network with cross-entropy loss.  However, we find that this mask prediction net does not work well for pose-varying human body.  The prediction tends to degrade in limbs parts as shown in Figure \ref{fig:Fig4_mask}.

To improve the performance of mask prediction for the pose-varying human body, we explore different methods and make two modifications.  The first is using the spatial transformer to predict binary mask instead of traditional EDN with softmax classifier which directly predicts the per-pixel labels.  The input of modified net is the binary mask of source image, and the coordinate flow is computed as intermediate tensor.  The final predicting result is not the per-pixel foreground probability like VSAP network, but a binary foreground mask.  Because of this, our mask prediction cannot be used to fuse multi-view synthesis result as it does not produce probability, but in return it predicts better mask for pose-varying humans.

\begin{figure*}
\begin{center}
\includegraphics[width=1\linewidth]{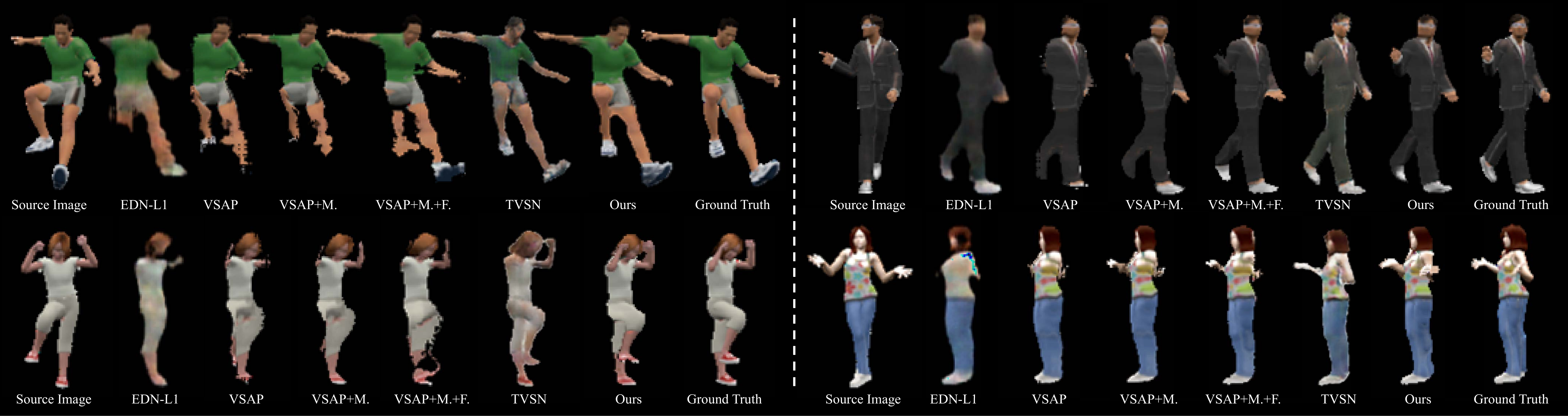}
\vspace{-2.5em}
\end{center}
   \caption{Comparison of low resolution result.}
\label{fig:Fig5_low_result}
\vspace{-1em}
\end{figure*}

The second modification is to take advantage of the predicted forward flow, which robustly reflects the structure in the front side.  Experimentally, we find that when the pose is complicated, the flow based transforming net often fails to maintain detailed structures and tends to produce over-smoothed mask margin, as shown in Figure \ref{fig:Fig4_mask} (c).  We propose to extract the mask of the transformed forward flow from the target mask, and merely predict the remaining part, namely residual mask.  In the training phase, the loss in the transformed flow region is masked out.  We denote the mask $M(x,y)$ as binary image, with $1$ as foreground and $0$ as background.  Given the transformed mask $M_{tran}(x,y)$ and target mask $M_{tgt}(x,y)$, the residual mask $M_R(x,y)$ is denoted as

\vspace{-0.2in}
\begin{equation}
M_R(x,y) = M_{tgt}(x,y)  \otimes  (\sim M_{tran}(x,y))
\end{equation}
\vspace{-0.2in}

where $\otimes$ is the per-pixel Boolean $and$ operation, and $\sim$ is the per-pixel Boolean $not$ operation.  In post-processing, we apply morphological close operation to eliminate tiny interval space in the residual mask. The final prediction mask $M_{final}(x,y)$ is generated by

\vspace{-0.2in}
\begin{equation}
M_{final}(x,y) = M_{pred}(x,y)  \oplus  M_{tran}(x,y)
\end{equation}
\vspace{-0.2in}

where $M_{pred}(x,y)$ is the mask produced by flow based transforming net.  $\oplus$ is per-pixel Boolean $or$ operation.

In our experiments, we tried two kinds of network input: the source mask together with transforming vector (denoted in VSAP) and single forward flow.  Both inputs achieve the similar mask prediction accuracy, so we choose the forward flow as the input to avoid introducing redundant transforming vectors.

\begin{table*}[t]
\centering
\caption{Quantitative comparison with previous methods. }
\label{tab:Tab1_comparison}
\begin{threeparttable}
    \begin{tabular}{c|cc|ccc|c}
        \hline
        \multirow{2}{*}{Method}            & \multicolumn{2}{c|}{image accuracy}             & \multicolumn{3}{c|}{flow accuracy}                                  & mask accuracy     \\ 
                                           & MSE $\downarrow$\tnote{1} & SSIM $\uparrow$   & MSE $\downarrow$ & $\delta_{1.25}$ $\uparrow$  & NCC $\uparrow$    & IoU $\uparrow$    \\ \hline
        EDN+L1 \cite{ECCV2016tatarchenko} & 96.83                       & 0.9488            & ---                         & ---               & ---               & 0.7528            \\ 
        TVSN\cite{CVPR2017park} & 85.35                       & 0.9519            & ---                         & ---               & ---               & 0.8344            \\ 
        VSAP \cite{ECCV2016zhou}          & 131.6                       & 0.9527            & 34.04                       & 0.5222            & 0.8093            & 0.7903            \\ 
        VSAP+M.                            & 90.44                       & 0.9596            & 34.04                       & 0.5222            & 0.8093            & 0.8907            \\ 
        VSAP+M.+F.                         & 87.34                       & 0.9617            & 13.01                       & 0.6288            & 0.8687            & 0.8831            \\ 
        Ours                                     & \textbf{72.86}            & \textbf{0.9670} & \textbf{2.207}            & \textbf{0.8630} & \textbf{0.9636} & \textbf{0.9109} \\ \hline
    \end{tabular}
    \begin{tablenotes}
        \footnotesize
        \item[1] For each metrics, $\uparrow$ means the larger the better and $\downarrow$ means the smaller the better.
    \end{tablenotes}
\end{threeparttable}
\vspace{-0.2in}
\end{table*}

\begin{figure*}
\begin{center}
\includegraphics[width=1\linewidth]{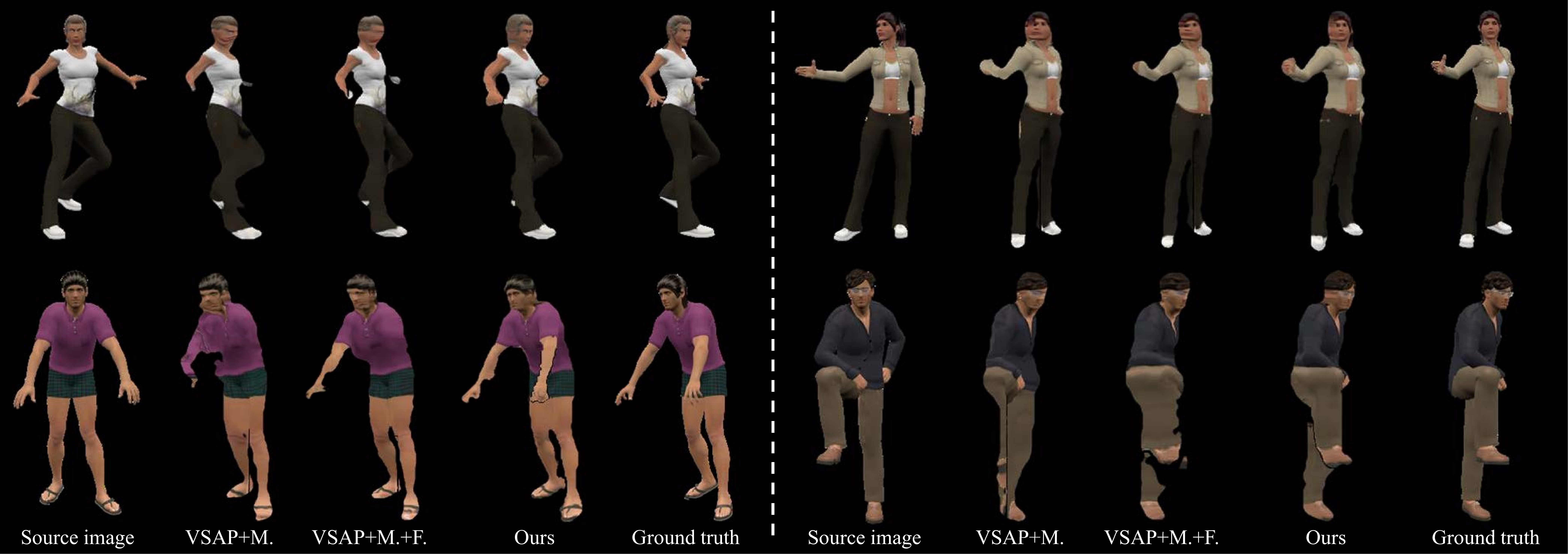}
\vspace{-2.5em}
\end{center}
   \caption{Comparison of high resolution result.  We recommend to zoom in the figure to see the detailed performance.}
\label{fig:Fig6_high_result}
\vspace{-0.8em}
\end{figure*}

\begin{figure*}
\begin{center}
\includegraphics[width=1\linewidth]{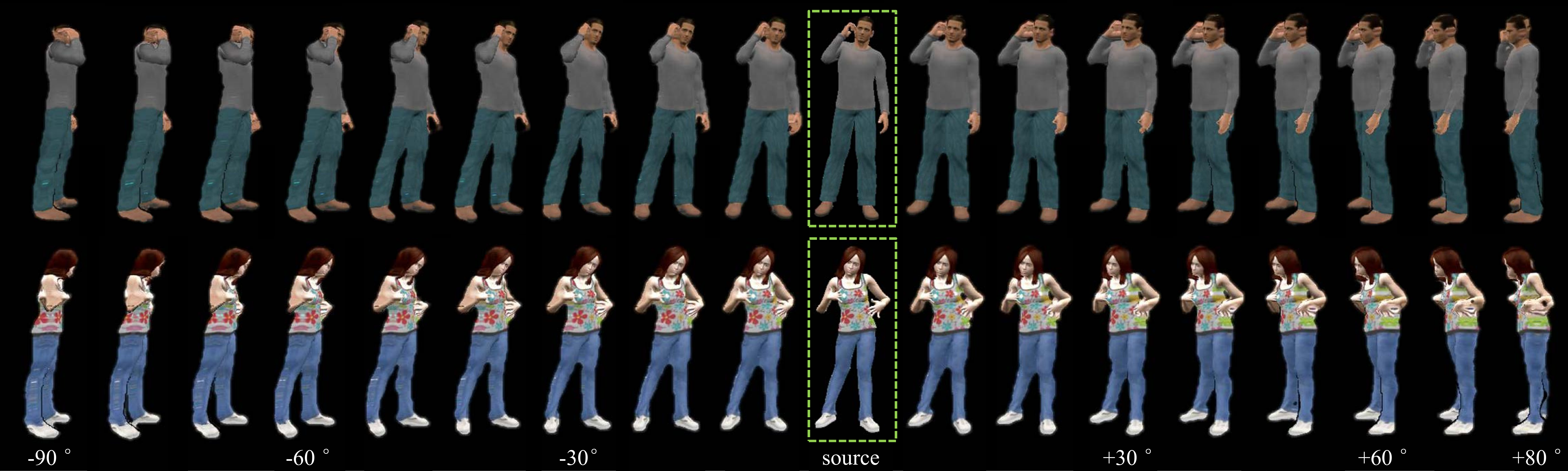}
\vspace{-2.5em}
\end{center}
   \caption{Results on synthesizing full viewpoints.}
\label{fig:Fig7_full_view}
\vspace{-1.2em}
\end{figure*}

\begin{figure*}
\begin{center}
\includegraphics[width=1\linewidth]{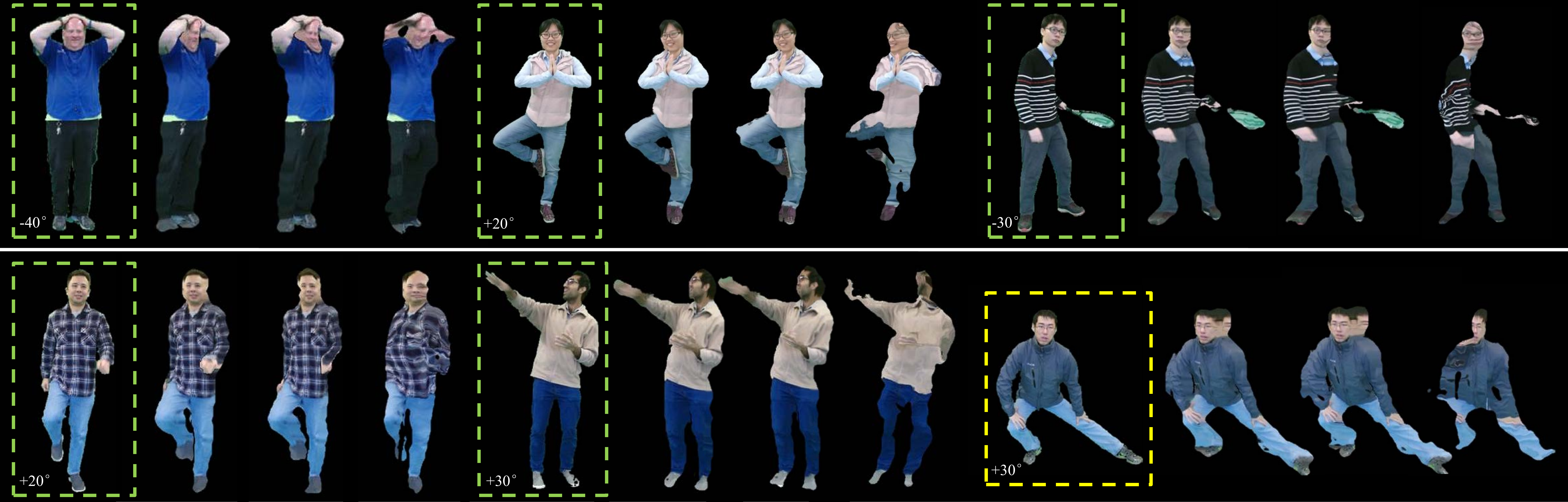}
\vspace{-2em}
\end{center}
   \caption{Synthesized results using real images. The image in the dashed box is the input image, and on its right there are (1) Our fine-tuned result; (2) Our result without fine-tuning; (3) VSAP+M. result.  The box with yellow dashed line indicates the failure case.}
\label{fig:Fig8_real}
\vspace{-1em}
\end{figure*}
\section{Experiments}
Since there is no publicly available dataset for a large number of 3D human body models (with different clothes and appearances), we create a synthetic dataset using the Poser software\footnote{http://my.smithmicro.com/poser-3d-animation-software.html}.  The access link to the dataset could be found at the end of Section \ref{sec:intro}.  Specifically, human models with 22 different appearances are generated and each of them is deformed into 200 to 1200 different poses, forming a dataset with over 10,000 human models.  We render the textured mesh of a human model with each specific pose to images, depth and flow respectively, and we select the front view as the source view, i.e. pose with $0^{\circ}$, while set other 17 views in the range of $[-90^{\circ}, 80^{\circ}]$ with interval $10^{\circ}$ as the target views.  
Each pair of source view and target view contains corresponding masks, depth images and ground truth backward flow from target view to source view.  
We randomly select $80\%$ of the pairs as training data and the rest as testing data.  

For training and evaluation, images, ground truth masks, depths and flow maps are rendered with a resolution of $200\times200$, which eases the learning of the networks.
After the model is trained, for visualization, we re-render a source image with a resolution of  $500\times500$, and upsample the estimated backward flow map to the same size to perform the view synthesis, yielding visually more satisfactory results.

\textbf{Training details.} In the training phase, the depth net is firstly trained, during which we augment the training data by randomly rotating the front view pose between $[-30^{\circ}, 30^{\circ}]$ with a interval of $5^{\circ}$.
Then, given the predicted depth results, the flow net and mask net are additionally trained.  
We use the `Adam' optimizer to train the three networks, with an initial learning rate as $1^{-4}$, and we reduce it by a ratio of $0.5$ at every 50,000 iterations. 
For flow net, the loss weight of image $\eta_y$ and flow $\eta_f$ in Equation \ref{equ:loss_func} are $10^{-6}$ and $1$ separately.

\subsection{Quantitative Results}
For evaluation, we randomly selected 2000 pairs from our testing data to compare different algorithms.
The metrics for evaluating synthesized images, backward flow and human mask are,
\begin{itemize}
\vspace{-0.1in}
\item Mean Squared Error(MSE),  which is used to measure color difference between synthesized image/flow and ground truth image/flow.  For backward flow, only the pixels with valid ground truth flow are counted.
\vspace{-0.1in}
\item Structural SIMilarity (SSIM) Index~\cite{TIP2004Wang, ICLR2016Mathieu}, which has value in $[-1,1]$, measuring the structural similarity between synthesized image and ground truth. 
\vspace{-0.1in}
\item Percentage of correctness under threshold $\delta$: Formally, for predicted flow $\overline{f_i}$ at pixel $i$, given ground truth $f_i$, it is regarded as correct if  $max(\frac{\overline{f_i}}{f_i},\frac{f_i}{\overline{f_i}})<\delta$ is satisfied. 
We count the portion of correctly predicted pixels. Here we set $\delta=1.25$.
\vspace{-0.1in}
\item Normalization Cross Correlation (NCC), which has value in the range of $[-1,1]$, measuring the correctness of backward flow direction.
\vspace{-0.1in}
\item Intersection over Union (IoU), which has value in the range of $[0,1]$, measuring the quality of segmented mask.

\end{itemize}
\vspace{-0.1in}
We compare our approach against several deep learning based state-of-the-art algorithms.  As introduced at Section \ref{sec:related},
EDN + L1 \cite{ECCV2016tatarchenko} directly synthesizes image pixels without intermediate representations using L1 loss.  
VSAP \cite{ECCV2016zhou} takes advantage of spatial transformer to produce intermediate backward flow for synthesis.  
TVSN~\cite{CVPR2017park} is a cascade system consists of a spatial transformer net (DOAFN) and a completion net.  The DOAFN has the same structure with VSAP, generating an appearance flow based synthesis result together with a visibility map.  Then the completion net refines it by hallucinating the invisible region.

We compare our method with previous methods in the quality of image, backward flow and segmented human mask. 
The comparison results are shown in Table \ref{tab:Tab1_comparison}. For evaluating the human segmentation mask, since EDN, TVSN does not have explicit mask produced, we use the non-zeros area from their predicted images.  

Though VSAP has proven its superiority towards EDN+L1 on rigid objects like chairs and cars\cite{ECCV2016zhou}, our experiments show that VSAP performs poorer than EDN+L1 on pose-varying human dataset.  The main problem is the poor mask prediction result, so we provide VSAP the masks from flow based mask prediction net, which makes it achieve lower MSE loss comparing to EDN+L1.  Additionally, we supplement the ground truth flow as supervision in training, denoted by `VSAP+M+F', yielding even better results from finding correct pixels at source images.  TVSN takes advantages of both adversarial and perceptual loss to improve on the base of VSAP, but its effectiveness is limited as it cannot improve the region where VSAP fails.  At the last row, our model produces the best results.  Through the flow evaluation we can see that our flow accuracy is markedly higher than VSAP+M.+F., which contributes a lot to prediction quality improvement.

We also compare our method with VSAP\cite{ECCV2016zhou} on rigid objects by following their experimental setup using the `car' and `chair' categories derived from ShapeNet\cite{ShapeNet}.  
Different from our setting for human body, the range of target views to synthesize are expanded to $[-180^{\circ}, 170^{\circ}]$ with an interval of $20^{\circ}$, and no front view assumption is applied.  
We keep the image size as $200\times200$ according to our model setup.  The quantitative comparison in Table \ref{tab:Tab2_rigid} shows that our method is comparable or slightly better than VSAP for rigid objects.  We believe it is because the cars and chair share similar shape that reduce the variation space, which alleviate the difficulty for a network to learn a good representation even without 3D geometry constraints.

\subsection{Visualization}
Figure \ref{fig:Fig5_low_result} and Figure \ref{fig:Fig6_high_result} visualize the prediction result of ours and other methods in low resolution $200\times200$ (LR) and high resolution $500\times500$ (HR) respectively.  
As indicated before, The LR result is directly generated from the network, while the HR prediction is warped from HR source image using an up-sampled backward flow with bi-cubic interpolation.  EDN+L1 and TVSN are not shown in the figure because these two methods don't produce the explicit flow for HR image synthesis.
As expected,  previous methods tend to have the wrong texture in parts of arms and legs, especially when self-occlusion occurs.  Our method does much better in tackling these issues as we leverage the geometry of the underlying human body.  
From the HR results, we can see more details on faces are well preserved, demonstrating the accuracy of the predicted flow.
Finally, Figure \ref{fig:Fig7_full_view} gives the prediction from a source view to all 17 target views, i.e. $-90^{\circ}$ to $+80^{\circ}$.

\begin{table}[t]
\centering
\caption{Comparison with rigid objects.}
\label{tab:Tab2_rigid}
\label{my-label}
\begin{tabular}{c|cc|cc}
\hline
\multirow{2}{*}{Method} & \multicolumn{2}{c|}{car} & \multicolumn{2}{c}{chair} \\ \cline{2-5}
                        & MSE             & SSIM            & MSE         & SSIM       \\ \hline
VSAP                    & 265.1          & 0.9061           & 499.3      & 0.8890      \\ 
Ours                    & 263.4          & 0.9058           & 464.3      & 0.8904           \\ \hline
\end{tabular}
\vspace{-1.5em}
\end{table}

\subsection{Model Transfer to Real Images.} 
Besides synthesized examples, we took many real world images using Kinect2 for testing the effectiveness of our model trained on synthetic data.  Here, we re-size the recorded images to fit the network input.
Specifically, we only fine-tune the depth network by 8,000 frames of RGB-D images for handling the domain transfer issue, while keeping the rest of the networks the same.
As shown in Figure \ref{fig:Fig8_real}, we compare the predictions with/without handling depth net domain transfer, and the prediction from 'VSAP+M.' which is presented before.  
The first observation is that depth net with domain transfer handled does improve the visual quality, while the model without fine-tuning can still maintain the pose structure. 
In contrast, VSAP is much more sensitive to unfamiliar appearances and shape from real data. This demonstrates that our model with 3D constraints mines more meaningful cues from the synthesized data, yielding better robustness.

\section{Conclusion}
We present a novel method that synthesizes novel views of the human body from a single image.  Previous methods like image generation network\cite{ECCV2016tatarchenko} and spatial transforming network\cite{ECCV2016zhou} are based on the assumption that the objects share similar shapes. The articulated and deformable human body renders previous methods ineffective.  We apply a strategy that first predicts shape from appearance, and then synthesizes the optical flow and mask.  A novel system architecture is developed, in which the flow and mask prediction networks follow a depth prediction network. The two networks are linked via a perspective projection layer in which geometric principles are explicitly applied. We show that our approach significantly improves the view synthesis quality for pose-varying human body.

Our method can still be improved in some aspects.  The lack of inference ability of our network makes our synthesis results are implausible when the rotation angle is larger than $90^{\circ}$.  Though TVSN\cite{CVPR2017park} explored this problem, we find that the proposed approach does not work well for pose-varying human data.  Besides, prior on human body may be used to further improve view synthesis quality. 

\vspace{-0.05in}
\section{Acknowledgements}
\vspace{-0.05in}
This work was supported in part by the US NSF grant IIP-1543172, a gift grant from Huawei,  China's National Natural Science Foundation (61627804, 61671236, 61332017),  and National Key R\&D project (grant no: 2017YFB1002803).

\clearpage

{\small
\bibliographystyle{ieee}
\bibliography{mybib}{}

\begin{thebibliography}{10}\itemsep=-1pt

\bibitem{SCAPE:sig05}
D.~Anguelov, P.~Srinivasan, D.~Koller, S.~Thrun, J.~Rodgers, and J.~Davis.
\newblock Scape: Shape completion and animation of people.
\newblock {\em Transactions on Graphics (ToG)}, 24(3):408--416, 2005.

\bibitem{Bogo:ECCV:2016}
F.~Bogo, A.~Kanazawa, C.~Lassner, P.~Gehler, J.~Romero, and M.~J. Black.
\newblock Keep it smpl: Automatic estimation of 3d human pose and shape from a
  single image.
\newblock In {\em European Conference on Computer Vision (ECCV)}, pages
  561--578, 2016.

\bibitem{TCSVT2017Cao}
Y.~Cao, Z.~Wu, and C.~Shen.
\newblock Estimating depth from monocular images as classification using deep
  fully convolutional residual networks.
\newblock {\em IEEE Transactions on Circuits and Systems for Video Technology
  (TCSVT)}, 2017.

\bibitem{ShapeNet}
A.~X. Chang, T.~Funkhouser, L.~Guibas, P.~Hanrahan, Q.~Huang, Z.~Li,
  S.~Savarese, M.~Savva, S.~Song, H.~Su, et~al.
\newblock Shapenet: An information-rich 3d model repository.
\newblock {\em arXiv preprint arXiv:1512.03012}, 2015.

\bibitem{ICCV2017Chen}
Q.~Chen and V.~Koltun.
\newblock Photographic image synthesis with cascaded refinement networks.
\newblock In {\em IEEE International Conference on Computer Vision (ICCV)},
  pages 1521--1529, 2017.

\bibitem{TPAMI2017Alex}
A.~Dosovitskiy, J.~T. Springenberg, M.~Tatarchenko, and T.~Brox.
\newblock Learning to generate chairs, tables and cars with convolutional
  networks.
\newblock {\em IEEE Transactions on Pattern Analysis and Machine Intelligence
  (TPAMI)}, 39(4):692--705, 2017.

\bibitem{NIPS2014Eigen}
D.~Eigen, C.~Puhrsch, and R.~Fergus.
\newblock Depth map prediction from a single image using a multi-scale deep
  network.
\newblock In {\em Advances in Neural Information Processing Systems (NIPS)},
  pages 2366--2374, 2014.

\bibitem{CVPR2017ZFan}
H.~Fan, H.~Su, and L.~Guibas.
\newblock A point set generation network for 3d object reconstruction from a
  single image.
\newblock In {\em IEEE Conference on Computer Vision and Pattern Recognition
  (CVPR)}, pages 2463--2471, 2017.

\bibitem{CVPR2015flynn}
J.~Flynn, I.~Neulander, J.~Philbin, and N.~Snavely.
\newblock Deepstereo: Learning to predict new views from the world's imagery.
\newblock In {\em IEEE Conference on Computer Vision and Pattern Recognition
  (CVPR)}, pages 5515--5524, 2016.

\bibitem{TPAMI2010Furukawa}
Y.~Furukawa and J.~Ponce.
\newblock Accurate, dense, and robust multiview stereopsis.
\newblock {\em IEEE Transactions on Pattern Analysis and Machine Intelligence},
  32(8):1362--1376, 2010.

\bibitem{ECCV2016Garg}
R.~Garg, G.~Carneiro, and I.~Reid.
\newblock Unsupervised cnn for single view depth estimation: Geometry to the
  rescue.
\newblock In {\em European Conference on Computer Vision (ECCV)}, pages
  740--756, 2016.

\bibitem{NIPS2014Goodfellow}
I.~Goodfellow, J.~Pouget-Abadie, M.~Mirza, B.~Xu, D.~Warde-Farley, S.~Ozair,
  A.~Courville, and Y.~Bengio.
\newblock Generative adversarial nets.
\newblock In {\em Advances in Neural Information Processing Systems (NIPS)},
  pages 2672--2680, 2014.

\bibitem{CVPR02016He}
K.~He, X.~Zhang, S.~Ren, and J.~Sun.
\newblock Deep residual learning for image recognition.
\newblock In {\em IEEE Conference on Computer Vision and Pattern Recognition
  (CVPR)}, pages 770--778, 2016.

\bibitem{ICANN2011Hinton}
G.~E. Hinton, A.~Krizhevsky, and S.~D. Wang.
\newblock Transforming auto-encoders.
\newblock In {\em International Conference on Artificial Neural Networks
  (ICANN)}, pages 44--51. Springer, 2011.

\bibitem{ICCV2017Huang}
R.~Huang, S.~Zhang, T.~Li, and R.~He.
\newblock Beyond face rotation: Global and local perception gan for
  photorealistic and identity preserving frontal view synthesis.
\newblock {\em IEEE International Conference on Computer Vision (ICCV)}, pages
  2439--2448, 2017.

\bibitem{NIPS2015jaderberg}
M.~Jaderberg, K.~Simonyan, A.~Zisserman, et~al.
\newblock Spatial transformer networks.
\newblock In {\em Advances in Neural Information Processing Systems (NIPS)},
  pages 2017--2025, 2015.

\bibitem{CVPR2017ji}
D.~Ji, J.~Kwon, M.~McFarland, and S.~Savarese.
\newblock Deep view morphing.
\newblock In {\em IEEE Conference on Computer Vision and Pattern Recognition
  (CVPR)}, pages 2155--2163, 2017.

\bibitem{ECCV2016Johnson}
J.~Johnson, A.~Alahi, and L.~Fei-Fei.
\newblock Perceptual losses for real-time style transfer and super-resolution.
\newblock In {\em European Conference on Computer Vision (ECCV)}, pages
  694--711. Springer, 2016.

\bibitem{TOG2016Kalantari}
N.~K. Kalantari, T.-C. Wang, and R.~Ramamoorthi.
\newblock Learning-based view synthesis for light field cameras.
\newblock {\em ACM Transactions on Graphics (TOG)}, 35(6):193, 2016.

\bibitem{TOG2014Kholgade}
N.~Kholgade, T.~Simon, A.~Efros, and Y.~Sheikh.
\newblock 3d object manipulation in a single photograph using stock 3d models.
\newblock {\em ACM Transactions on Graphics (TOG)}, 33(4):127, 2014.

\bibitem{3DV2016Laina}
I.~Laina, C.~Rupprecht, V.~Belagiannis, F.~Tombari, and N.~Navab.
\newblock Deeper depth prediction with fully convolutional residual networks.
\newblock In {\em International Conference on 3D Vision (3DV)}, pages 239--248,
  2016.

\bibitem{ICRA2017Liao}
Y.~Liao, L.~Huang, Y.~Wang, S.~Kodagoda, Y.~Yu, and Y.~Liu.
\newblock Parse geometry from a line: Monocular depth estimation with partial
  laser observation.
\newblock In {\em IEEE International Conference on Robotics and Automation
  (ICRA)}, pages 5059--5066, 2017.

\bibitem{CVPR2019Liu}
Y.~Liu, X.~Cao, Q.~Dai, and W.~Xu.
\newblock Continuous depth estimation for multi-view stereo.
\newblock In {\em IEEE Conference on Computer Vision and Pattern Recognition
  (CVPR)}, pages 2121--2128, 2009.

\bibitem{ICRA2018Ma}
F.~Ma and S.~Karaman.
\newblock Sparse-to-dense: depth prediction from sparse depth samples and a
  single image.
\newblock {\em IEEE International Conference on Robotics and Automation
  (ICRA)}, 2018.

\bibitem{NIPS2017Mao}
X.-J. Mao, C.~Shen, and Y.-B. Yang.
\newblock Image restoration using convolutional auto-encoders with symmetric
  skip connections.
\newblock In {\em Advances in Neural Information Processing Systems (NIPS)},
  pages 2810--2818, 2016.

\bibitem{ICLR2016Mathieu}
M.~Mathieu, C.~Couprie, and Y.~LeCun.
\newblock Deep multi-scale video prediction beyond mean square error.
\newblock In {\em International Conference on Learning Representation (ICLR)},
  2016.

\bibitem{SCIENCE1971NShepard}
R.~N.~Shepard and J.~Metzler.
\newblock Mental rotation of three-dimensional objects.
\newblock {\em Science (New York, N.Y.)}, 171:701--3, 03 1971.

\bibitem{CVPR2017park}
E.~Park, J.~Yang, E.~Yumer, D.~Ceylan, and A.~C. Berg.
\newblock Transformation-grounded image generation network for novel 3d view
  synthesis.
\newblock In {\em IEEE Conference on Computer Vision and Pattern Recognition
  (CVPR)}, pages 702--711, 2017.

\bibitem{TPAMI2017Rematas}
K.~Rematas, C.~H. Nguyen, T.~Ritschel, M.~Fritz, and T.~Tuytelaars.
\newblock Novel views of objects from a single image.
\newblock {\em IEEE Transactions on Pattern Analysis and Machine Intelligence
  (TPAMI)}, 39(8):1576--1590, 2017.

\bibitem{IJCV2008Saxena}
A.~Saxena, S.~H. Chung, and A.~Y. Ng.
\newblock 3d depth reconstruction from a single still image.
\newblock {\em International Journal of Computer Vision (IJCV)}, 76(1):53--69,
  2008.

\bibitem{CVPR2006Seitz}
S.~M. Seitz, B.~Curless, J.~Diebel, D.~Scharstein, and R.~Szeliski.
\newblock A comparison and evaluation of multi-view stereo reconstruction
  algorithms.
\newblock In {\em IEEE Conference on Computer vision and pattern recognition
  (CVPR)}, pages 519--528, 2006.

\bibitem{Sigal:nips:2008}
B.~A. Sigal, L. and M.~Black.
\newblock Combined discriminative and generative articulated pose and non-rigid
  shape estimation.
\newblock In {\em Advances in Neural Information Processing Systems (NIPS)},
  pages 1337–--1344, 2008.

\bibitem{ICCV2017Srinivasan}
P.~P. Srinivasan, T.~Wang, A.~Sreelal, R.~Ramamoorthi, and R.~Ng.
\newblock Learning to synthesize a 4d rgbd light field from a single image.
\newblock {\em IEEE International Conference on Computer Vision (ICCV)}, pages
  2262--2270, 2017.

\bibitem{SIGGRAPH2014Su}
H.~Su, Q.~Huang, N.~J. Mitra, Y.~Li, and L.~Guibas.
\newblock Estimating image depth using shape collections.
\newblock {\em Transactions on Graphics (ToG)}, 33(4):37, 2014.

\bibitem{ICCV2015Su}
H.~Su, F.~Wang, L.~Yi, and L.~Guibas.
\newblock 3d-assisted image feature synthesis for novel views of an object.
\newblock {\em IEEE International Conference on Computer Vision (ICCV)}, pages
  2677--2685, 2015.

\bibitem{ECCV2016tatarchenko}
M.~Tatarchenko, A.~Dosovitskiy, and T.~Brox.
\newblock Multi-view 3d models from single images with a convolutional network.
\newblock In {\em European Conference on Computer Vision (ECCV)}, pages
  322--337, 2016.

\bibitem{TIP2004Wang}
Z.~Wang, A.~C. Bovik, H.~R. Sheikh, and E.~P. Simoncelli.
\newblock Image quality assessment: From error visibility to structural
  similarity.
\newblock {\em IEEE Transactions on Image Processing (TIP)}, 13(4):600--612,
  2004.

\bibitem{NIPS2015Yang}
J.~Yang, S.~E. Reed, M.-H. Yang, and H.~Lee.
\newblock Weakly-supervised disentangling with recurrent transformations for 3d
  view synthesis.
\newblock In {\em Advances in Neural Information Processing Systems (NIPS)},
  pages 1099--1107, 2015.

\bibitem{arXiv2017Zhao}
B.~Zhao, X.~Wu, Z.-Q. Cheng, H.~Liu, and J.~Feng.
\newblock Multi-view image generation from a single-view.
\newblock {\em arXiv preprint arXiv:1704.04886}, 2017.

\bibitem{ECCV2016zhou}
T.~Zhou, S.~Tulsiani, W.~Sun, J.~Malik, and A.~A. Efros.
\newblock View synthesis by appearance flow.
\newblock In {\em European Conference on Computer Vision (ECCV)}, pages
  286--301, 2016.

\bibitem{TCSVT2017Zhu}
H.~Zhu, Y.~Liu, J.~Fan, Q.~Dai, and X.~Cao.
\newblock Video-based outdoor human reconstruction.
\newblock {\em IEEE Transactions on Circuits and Systems for Video Technology},
  27(4):760--770, 2017.

\bibitem{FCS2017Zhu}
H.~Zhu, Y.~Nie, T.~Yue, and X.~Cao.
\newblock The role of prior in image based 3d modeling: a survey.
\newblock {\em Frontiers of Computer Science}, 11(2):175--191, 2017.

\end{thebibliography}
}

\end{document}